\documentclass{article} 
\usepackage{iclr2026_conference,times}

\usepackage{amsmath,amsfonts,bm}

\def\eqref#1{equation~\ref{#1}}

\def\1{\bm{1}}

\DeclareMathAlphabet{\mathsfit}{\encodingdefault}{\sfdefault}{m}{sl}
\SetMathAlphabet{\mathsfit}{bold}{\encodingdefault}{\sfdefault}{bx}{n}

\usepackage{hyperref}
\usepackage{url}
\usepackage{todonotes}
\usepackage{multirow}
\usepackage{multicol}
\usepackage{caption}
\usepackage{todonotes}
\DeclareUnicodeCharacter{200E}{}

\title{DiffuSAM: Diffusion Guided Zero-Shot Object Grounding for Remote Sensing Imagery}

\author{Geet Sethi \\
Indian Institute of Technology Bombay \\
\texttt{23b2258@iitb.ac.in}
\And
Panav Shah \\
Indian Institute of Technology Bombay \\
\texttt{panav.shah@iitb.ac.in}
\AND
Ashutosh Gandhe \\
Indian Institute of Technology Bombay \\
\texttt{22b2409@iitb.ac.in‎ ‎ ‎ ‎ ‎ ‎ ‎ ‎ ‎ ‎ ‎ ‎ ‎ ‎ ‎ ‎ ‎ ‎ ‎‎‎}
\And
Soumitra Nayak \\
Indian Institute of Technology Bombay \\
\texttt{22b0984@iitb.ac.in}
}

\iclrfinalcopy 
\begin{document}

\maketitle

\begin{abstract}
    
Diffusion models have emerged as powerful tools for a wide range of vision tasks, including text-guided image generation and editing. In this work, we explore their potential for object grounding in remote sensing imagery. We propose a hybrid pipeline that integrates diffusion-based localization cues with state-of-the-art segmentation models such as RemoteSAM and SAM3 to obtain more accurate bounding boxes. By leveraging the complementary strengths of generative diffusion models and foundational segmentation models, our approach enables robust and adaptive object localization across complex scenes. Experiments demonstrate that our pipeline significantly improves localization performance, achieving over a 14\% increase in Acc@0.5 compared to existing state-of-the-art methods.
\end{abstract}

\section{Introduction}
Remote sensing focuses on analyzing satellite and aerial imagery to identify and localize objects such as buildings, vehicles, roads, and natural features. Accurate object localization in this domain is critical for applications including urban planning, disaster response, environmental monitoring, and infrastructure analysis. 

Developing specialized object detection models for such tasks typically requires large-scale annotated datasets, which are expensive and difficult to obtain. As a result, there is a need for data-efficient approaches that can perform accurate grounding without requiring large-scale task-specific training. Furthermore, models trained on specific datasets often struggle to generalize across different geographic regions or imaging conditions. Recent advances in foundation models have introduced promising alternatives. Segmentation models such as SAM3 \citep{carion2025sam} enable strong zero-shot localization capabilities, while diffusion models \citep{ho2020denoisingdiffusionprobabilisticmodels} provide powerful text-guided image generation and editing. In addition, RemoteSAM \citep{yao2025remotesam}, a segmentation model specifically trained on satellite imagery, has emerged as the current state-of-the-art for remote sensing visual grounding. These developments collectively offer an opportunity to design flexible object grounding pipelines that combine general-purpose foundation models with domain-specialized models, without relying on extensive task-specific training.

In this work, we propose DiffuSAM, a hybrid pipeline that combines diffusion-based visual grounding with state-of-the-art segmentation models to obtain accurate bounding boxes in remote sensing imagery. Our approach leverages diffusion models to generate approximate localization cues from natural language prompts and refines them using SAM-based segmentation. By adaptively selecting between RemoteSAM and SAM3 based on the characteristics of the region of interest, the method effectively handles both complex multi-object scenes and precise small-object localization. Experiments demonstrate that this approach improves the localization accuracy by more than 14\% Acc@0.5\footnote{Acc@0.5 denotes accuracy at IoU $>$ 0.5.} over the baseline methods.

\section{Methodology}

\begin{figure}[t]
    \centering
    \includegraphics[width=0.95\linewidth]{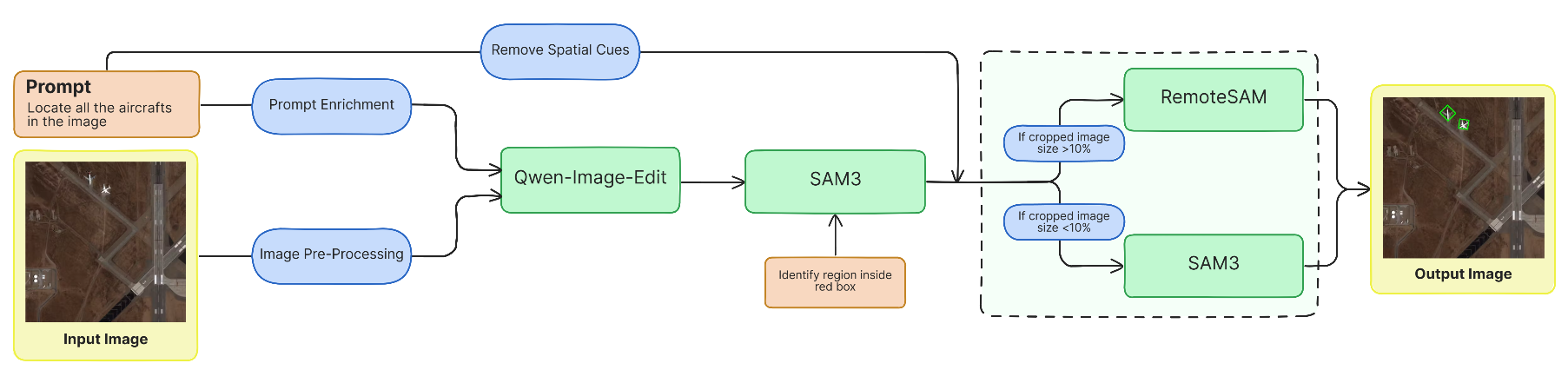}
    \caption{DiffuSAM Pipeline}
    \label{fig:pipeline}
\end{figure}

\subsection{Overview}
The proposed DiffuSAM pipeline integrates diffusion-based image editing with foundational segmentation models to perform text-guided object localization. Given an input image and a textual description of the target object, the pipeline generates an approximate region of interest and iteratively refines it to obtain an accurate bounding box.

\subsection{Pre-processing}

Satellite imagery is often affected by haze, low contrast, and reduced visibility due to atmospheric conditions such as pollution, humidity, and sensor limitations. These factors can degrade image quality and make it difficult for models trained on clean, natural images to reliably recognize objects. To mitigate this issue, we incorporate a preprocessing step to enhance visual clarity before applying the grounding pipeline.

The input image is first converted from the RGB color space to the CIELAB color space \citep{597279} to separate luminance information from chromatic components. Contrast-Limited Adaptive Histogram Equalization (CLAHE) \citep{Zuiderveld1994ContrastLA} is then applied to the luminance channel to enhance local contrast and reduce haze while preventing excessive noise amplification. The enhanced luminance channel is subsequently recombined with the original chromatic channels and converted back to RGB. Finally, an unsharp masking operation is performed to further improve edge sharpness and overall visual clarity.

Apart from the image, the input natural language query is also rewritten using Qwen3-8b \citep{yang2025qwen3technicalreport}, a small LLM, to remove unnecessary instructions from the query which tend to misguide the model leading to incorrect bounding boxes.

\subsection{Diffusion-Based Localization}

We next employ Qwen-Image-Edit \citep{yin2025qwenimagelayered}, a diffusion-based image editing model, to obtain an approximate localization of the target object. Given an input image and a textual query, the model is prompted to generate an edited version of the image in which the described object is highlighted using red bounding boxes. This process effectively converts a natural language description into a coarse spatial cue, providing an initial estimate of the object’s location without requiring any task-specific training or manually annotated remote sensing data.

\subsection{Initial Segmentation with SAM3}
The edited image containing the red bounding boxes is passed to SAM3 \citep{carion2025sam}, which generates segmentation masks constrained to the highlighted regions. These masks are used to derive initial approximate bounding boxes for the objects of interest. This step transforms the coarse, diffusion-generated visual cues into structured spatial representations that can be further refined in subsequent stages.

\subsection{Adaptive Refinement Strategy}
At this stage, the textual prompt is refined by removing directional keywords such as ``left'' or ``right'' using a simple string replacement procedure. Since spatial relationships are already implicitly encoded in the cropped image, retaining such terms can introduce ambiguity and negatively affect segmentation performance.

To further refine the initial localization, the image region corresponding to each predicted bounding box is cropped and processed independently. Based on the relative size of the cropped region, the pipeline dynamically selects from two segmentation models with complementary strengths:

\begin{itemize}
    \item RemoteSAM \citep{yao2025remotesam} is used when the cropped region occupies more than $p\%$ of the original image. Larger regions often contain multiple objects or significant background context, and RemoteSAM is better suited for such scenarios due to its training on visual grounding tasks in satellite imagery.
    \item SAM3 \citep{carion2025sam} is used when the cropped region is small. In these cases, the initial localization is typically more precise, and SAM3 provides more accurate fine-grained segmentation.
\end{itemize}

The parameter $p$ is treated as a hyperparameter controlling the model selection strategy. Empirically, we found that setting $p = 10$ yields the best overall performance across datasets.

\subsection{Final Bounding Box Extraction}
The final mask produced by either RemoteSAM or SAM3 is used to compute the definitive bounding box around the target object. This bounding box serves as the output of the pipeline. Figure~\ref{fig:pipeline} illustrates the complete architecture of the proposed method.

To improve robustness, a fallback mechanism is incorporated. If the selected model fails to produce a valid segmentation mask, the alternative model is automatically invoked. If both models are unable to generate a mask, the system defaults to using the bounding box produced by the diffusion model as the final output.

\begin{figure}[tt]
    \centering
    \begin{minipage}{0.48\linewidth}
        \centering
        \includegraphics[width=\linewidth]{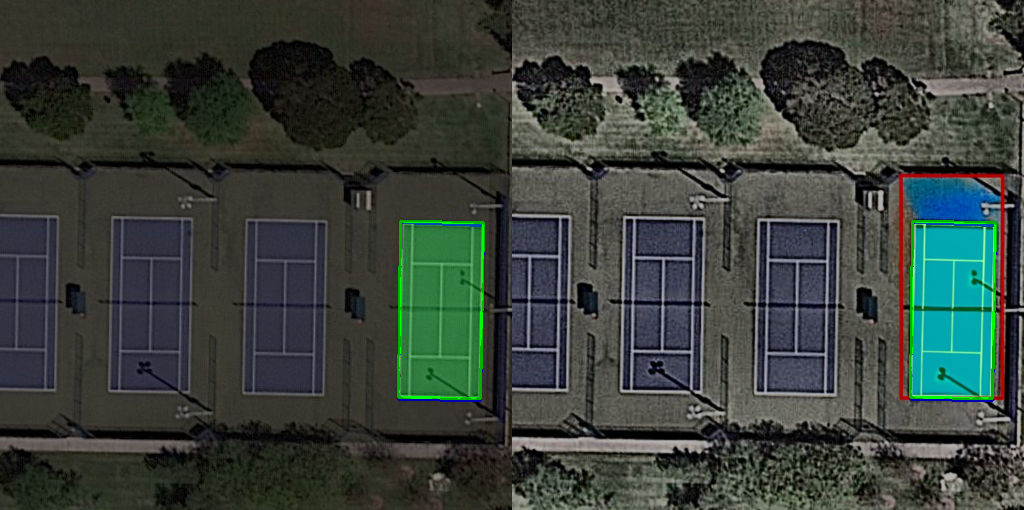}
        \caption{Successful localization by DiffuSAM for the prompt: ``The tennis court located on the right side of the image with a blue playing surface.''}
        \label{fig:tennis}
    \end{minipage}\hfill
    \begin{minipage}{0.48\linewidth}
        \centering
        \includegraphics[width=\linewidth]{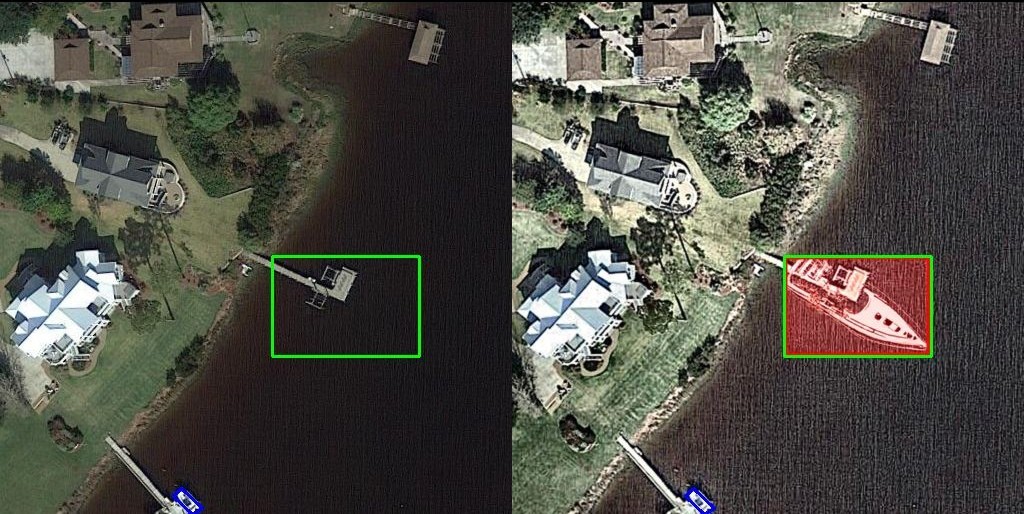}
        \caption{Failure case showing hallucinated localization by DiffuSAM for the prompt: ``The unique ship positioned in the lower central region of the image.''}
        \label{fig:hallucination}
    \end{minipage}
    \caption*{Red box: initial bounding box generated by the diffusion model; Green box: final prediction produced by the pipeline; Blue box: ground-truth annotation.}
\end{figure}

\section{Experiments}
\subsection{Datasets}
\begin{itemize}
    \item VRS Bench \citep{li2024vrsbench}: A large-scale vision–language benchmark for remote sensing supporting captioning, VQA, and visual grounding. It contains 29,614 images at $512 \times 512$ resolution, along with 52,472 referring-expression annotations. The dataset provides high-quality, human-verified annotations generated through a semi-automatic pipeline, making it suitable for evaluating text-guided localization.

    \item NWPU-VHR-10 \citep{Cheng_2016}: A widely used very-high-resolution remote sensing dataset for object detection. It consists of 800 images with 3,651 annotated objects across 10 categories, including vehicles, ships, and buildings. All instances are annotated with axis-aligned bounding boxes, enabling evaluation of localization in complex scenes.
\end{itemize}

\subsection{Results and Comparison}
We evaluate the proposed DiffuSAM pipeline on both VRSBench \citep{li2024vrsbench} and NWPU-VHR-10 \citep{Cheng_2016}, and compare it with existing state-of-the-art grounding and segmentation methods, including RemoteSAM \citep{yao2025remotesam}, SAM3 \citep{carion2025sam}, EarthMind \citep{shu2025earthmindleveragingcrosssensordata}, and Falcon \citep{yao2025falcon}. On VRSBench, our method performs better than the current best model, RemoteSAM, with a marginal increase in Acc@0.5. On the NWPU-VHR-10 dataset as well, DiffuSAM outperforms RemoteSAM by 14\% in Acc@0.5, demonstrating the effectiveness of our hybrid approach in challenging high-resolution scenarios.

The difference in performance across the two datasets can be largely attributed to annotation formats. NWPU-VHR-10 provides axis-aligned bounding boxes, however VRSBench provides oriented bounding boxes, which leads to a small evaluation mismatch.

Table~\ref{tab:results} presents a quantitative comparison of DiffuSAM with RemoteSAM, SAM3, EarthMind, and Falcon. The results highlight that our approach achieves competitive or superior performance without requiring task-specific training, demonstrating the advantage of combining diffusion models with foundation segmentation models for robust object grounding.

\begin{table*}[h]
\centering
\begin{tabular}{l | c c c | c c c}
\hline
\multirow{2}{*}{\textbf{Model}} 
& \multicolumn{3}{c}{\textbf{VRSBench}} 
& \multicolumn{3}{c}{\textbf{NWPU-VHR-10}} \\
\cline{2-7}
& \textbf{mIoU} & \textbf{Acc@0.5} & \textbf{\nolinkurl{Acc@0.7}}
& \textbf{mIoU} & \textbf{Acc@0.5} & \textbf{\nolinkurl{Acc@0.7}} \\
\hline
RemoteSAM & 0.6283 & 0.7812 & 0.3931 & 0.5662 & 0.6624 & 0.2484 \\
SAM3      & 0.1635 & 0.2051 & 0.0887 & 0.5009 & 0.6352 & 0.1683 \\
EarthMind & 0.5961 & 0.7414 & 0.3732 & 0.5256 & 0.6088 & 0.1844 \\
Falcon    & 0.5802 & 0.7221 & \textbf{0.4709} & 0.5415 & 0.6271 & 0.3051 \\
DiffuSAM  & \textbf{0.6414} & \textbf{0.7913} & 0.4051 & \textbf{0.6180} & \textbf{0.8049} & \textbf{0.3383} \\
\hline
\end{tabular}
\caption{Evaluation of different models on VRSBench and NWPU-VHR-10 datasets.}
\label{tab:results}
\end{table*}

\section{Limitations}

A primary limitation of our approach stems from the generative nature of diffusion models. When the target object is difficult to locate or absent, Qwen-Image-Edit \citep{yin2025qwenimagelayered} may hallucinate (Figure~\ref{fig:hallucination}) and generate a synthetic object rather than accurately highlighting the correct region. 

Additionally, the initial bounding box from the diffusion model is axis-aligned, which may include unnecessary objects in the selected region of interest and interfere with later pipeline stages.


\section{Discussion and Future Work}
In this work, we presented DiffuSAM, a hybrid pipeline that combines diffusion-based visual grounding with foundational segmentation models for object localization in remote sensing imagery. By leveraging diffusion models to generate approximate localization cues and refining them using SAM3 and RemoteSAM, the proposed method enables accurate and data-efficient bounding box prediction without requiring task-specific training. Experimental results on standard benchmarks demonstrate that DiffuSAM achieves significant improvements over baseline approaches. These results highlight the potential of integrating generative diffusion models with domain-specialized segmentation models for robust remote sensing applications. 

Future work includes controlling the diffusion model to reduce hallucinations and use generative models fine-tuned on remote sensing datasets to remove haziness and other obstructions from the satellite images to further improve the outputs from the pipeline.

\newpage

\bibliography{iclr2026_conference}
\bibliographystyle{iclr2026_conference}

\newpage

\end{document}